\documentclass[times, preprint, 10pt]{elsarticle}
\usepackage{array}
\usepackage[utf8]{inputenc} 
\usepackage[T1]{fontenc}    
\usepackage{hyperref}       
\usepackage{url}            
\usepackage{booktabs}       
\usepackage{amsfonts}       
\usepackage{nicefrac}       
\usepackage{microtype}      
\usepackage{xcolor}         
\usepackage{graphicx}
\usepackage{amsmath}
\usepackage{wrapfig}
\usepackage{color}
\usepackage{multirow}
\usepackage{comment}
\usepackage{caption}
\usepackage{subcaption}
\usepackage{pifont}
\usepackage{fancyhdr} 
\definecolor{myyellow}{RGB}{190,144,0}
\definecolor{mygreen}{RGB}{0,136,51}
\definecolor{myblue}{RGB}{0,102,204}

\title{DePatch: Towards Robust Adversarial Patch for Evading Person Detectors in the Real World}
\author[1]{Jikang Cheng} 
\author[2]{Ying Zhang}
\author[1]{Zhongyuan Wang\corref{cor1}}
\author[1]{Zou Qin}
\author[2]{Chen Li}
\address[1]{National Engineering Research Center for Multimedia Software, School of Computer Science, Wuhan University, Wuhan, China}
\address[2]{WeChat, Tencent, Shenzhen, China}

\cortext[cor1]{Corresponding author.}

\begin{document}
\begin{abstract}
Recent years have seen an increasing interest in physical adversarial attacks, which aim to craft deployable patterns for deceiving deep neural networks, especially for person detectors. However, the adversarial patterns of existing patch-based attacks heavily suffer from the \textbf{self-coupling} issue, where a degradation, caused by physical transformations, in any small patch segment can result in a complete adversarial dysfunction, leading to poor robustness in the complex real world. Upon this observation, we introduce the Decoupled adversarial Patch (DePatch) attack to address the self-coupling issue of adversarial patches. Specifically, we divide the adversarial patch into block-wise segments, and reduce the inter-dependency among these segments through randomly erasing out some segments during the optimization. 
We further introduce a border shifting operation and a progressive decoupling strategy to improve the overall attack capabilities.
Extensive experiments demonstrate the superior performance of our method over other physical adversarial attacks, especially in the real world.
\end{abstract}
\begin {keyword}
Adversarial Example, Physical Attack, Person Detector
\end {keyword}
\maketitle

\section{Introduction}
The vulnerability of Deep Neural Networks (DNNs) under adversarial attacks \cite{adv-digi1} has received significant attention in recent years. Notably, some physical adversarial examples can even be constructed in real-world scenarios to trick DNNs into producing erroneous outcomes. Given that the \textit{Person} class plays a critical role in various applications of object detection, the importance of research on physical attacks targeting person detection cannot be overstated. Consequently, several patch-based attack methods \cite{advPatch,advShirt,advShirt2,advTexture} have been proposed to evade person detectors \cite{yolov2,yolov3,retinaNet,faster-rcnn,mask-rcnn}. These methods aim to generate adversarial posters or craft adversarial clothes that make a person invisible to object detectors in the real world. 

\begin{figure}

	\centering
	\includegraphics[width=\textwidth]{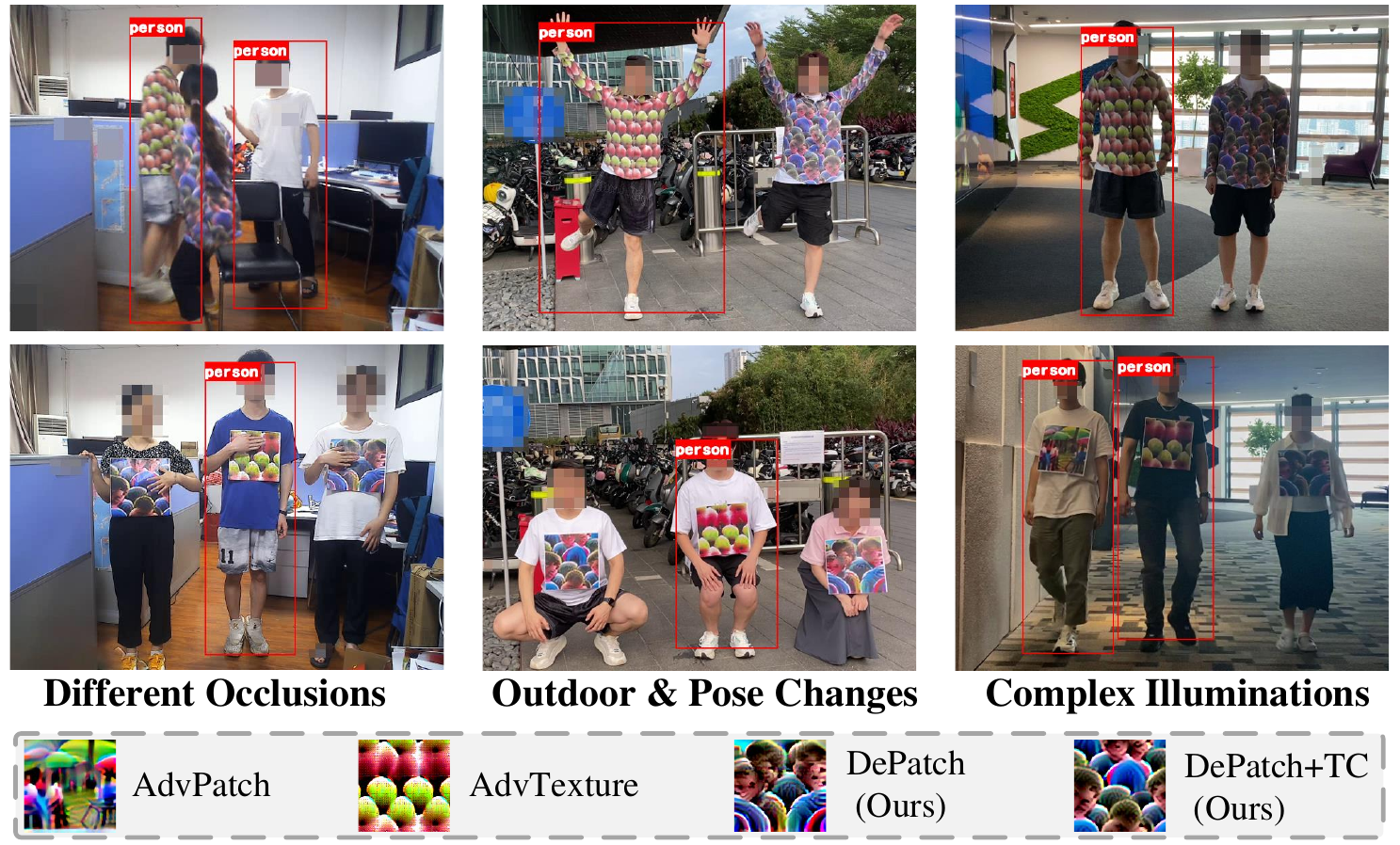}
	\caption{Attacking YOLOv2 \cite{yolov2} using different adversarial patches in the real world. Our method remains effective in multiple complex conditions. The mosaics are added after detection for privacy.}
	\label{fig:main-visual}
\end{figure}

However, while existing methods demonstrate promising adversarial effectiveness in the ideal digital world, their robustness against various degradations in the complex real world remains deficient. In this paper, we reveal the self-coupling issue of existing patch-based attacks, which severely limits the robustness of adversarial patches. In particular, existing patch-based methods optimize all segments of the entire patch simultaneously during training, and the adversarial effectiveness of each segment within the patch is dependent on other segments in the same patch. In other words, a degradation in any small segment of the patch could easily lead to a failure attack. As shown in Figure~\ref{fig:main-visual}, when the patch is deteriorated by partial occlusions, pose changes, outdoor, or illuminations, which frequently occur in real-world scenarios, previous methods like AdvPatch \cite{advPatch} and AdvTexture \cite{advTexture} lose the capability of hiding targets from the person detector.

To address the patch self-coupling issue, we propose a novel patch-based method named \textbf{Decoupled adversarial Patch (DePatch)} attack. The proposed DePatch attack aims to decouple the inter-dependency among all possible segments within the patch during optimization. Specifically, we perform patch decoupling by stochastically disabling a certain ratio of patch pixels during optimization. To guarantee the decoupling of any segment within the patch from all other segments, we erase out segments in a block-wise manner and employ border shifting. Unlike the widely-adopted data augmentation methods~\cite{cutout,gridmask}, the nature of attaching patches to images allows us to make the erased segments \textit{transparent} instead of setting to fixed pixel values (\textit{e.g.}, black, white or noise). 
This helps us avoid the issue of fixed-value regions affecting network training, which impedes segments from independently exhibiting adversarial effectiveness.
Subsequently, we propose a Progressive Decoupling Strategy (PDS) to progressively adjust block-wise decoupling variants of the decouple granularity and decouple ratio, which helps to address diverse degradations and strengthen the overall attack capabilities.
Meanwhile, the TC technique \cite{advTexture} can be incorporated into the DePatch optimization process, which allows DePatch to be expandable and can be used for clothing-based applications in the real world. As shown in Figure~\ref{fig:main-visual}, our proposed DePatch achieves good performance in attacking YOLOv2 \cite{yolov2} under various complex scenarios, such as occlusions, outdoor scenes, pose changes, and complex illuminations.

Our contributions are summarized as follows:
\begin{itemize}
	\item  We reveal the patch self-coupling issue for patch robustness in existing patch-based attacks, which inherently obstacles the real-world applications involving complex physical conditions.
	\item We propose the Decoupled adversarial Patch (DePatch) attack to address the self-coupling issue. Specifically, we conduct block-wise decoupling with the border shifting during training and a variant-adjusting strategy tailored for block-wise decoupling to endow individual patch segments with effectiveness.
	\item Extensive experiments demonstrate the superior robustness of the proposed DePatch against various physical degradations in both the digital and real world.
\end{itemize}

\section{Related Work}
\subsection{Physical Adversarial Example}
Since Szegedy \textit{et al.} \cite{adv-first} first pointed out that adversarial examples can cause DNNs to make mistakes, numerous adversarial attack methods have been proposed. In addition to digital attacks \cite{adv-digi1,pgd,FGSM,adv-first-2,adv-pr3}, physical adversarial examples \cite{phy1,deformable,PS-GAN,shadows,laser,stealthy3,adv-pr1,adv-pr2} that can be applied in real-world scenarios have received increasing attention. Physical adversarial examples can be improved in two primary directions: Effectiveness and stealthiness. Improving effectiveness aims to achieve more robust and effective adversarial attacks in real-world scenarios \cite{deformable,PS-GAN}, while improving stealthiness is intended to make adversarial examples more imperceptible to the human eyes \cite{laser,stealthy3,shadows}. In this paper, we mainly focus on the effectiveness of the physical adversarial examples.

\subsection{Patch-based Physical Attack for Person Detectors}
Patch-based physical attack \cite{patch-first,upc,dta,nps} is one of the most practical and effective approaches in physical attacks, and it has been widely applied to evade object detectors \cite{advPatch,advShirt,advShirt2,advTexture,upc,dta,tsea}. For person detection, Thys \textit{et al.} \cite{advPatch} attempted to simulate some simple physical transformations (\textit{i.e.}, rotations, scale, noises, brightness, and contrast) while training the adversarial patch according to Expectation over Transformation (EoT) \cite{eot}. They also introduced the Non Printability Score (NPS) \cite{nps} that encourages the pixel values of the generated patch closer to the colors that can be printed on cardboard. Xu \textit{et al.} \cite{advShirt} and Wu \textit{et al.} \cite{advShirt2} both considered printing adversarial patches on clothing and conducted extensive physical experiments. Xu \textit{et al.} \cite{advShirt} further proposed using Thin Plate Spline (TPS) to simulate the non-rigid deformation due to a moving person's posture changes. Hu \textit{et al.} \cite{advTexture} proposed the Toroidal-Cropping-based Expandable Generative Attack (TC-EGA), which crafts the adversarial patch with repetitive structures through a two-stage optimization process. The generated patch can be expanded to cover the entire body, thus deceiving the detectors from multiple viewing angles. These methods \cite{advPatch,advShirt,advShirt2,advTexture} have made particular improvements by simulating different real-world scenarios. Huang \textit{et al.}~\cite{tsea} propose a Transfer-based Self-Ensemble Attack (T-SEA) to tackle the transferability issue in the digital space.
However, their performance significantly deteriorates under simulated physical transformations within the digital world. Consequently, in real-world conditions, these methods encounter further difficulties in sustaining their robustness due to the variability in physical conditions.

\subsection{Data Augmentation via Information Deletion}~\label{sec:aug}

The proposed decoupling operation shares a formal resemblance to data augmentation via information deletion \cite{HaS, cutout, randomerasing, gridmask}. Singh \textit{et al.} introduced Hide-and-Seek (HaS) \cite{HaS}, which involves dividing training set images into segments and partially hiding them. Both Cutout \cite{cutout} and Random Erasing \cite{randomerasing} are proposed to delete pixels within randomly sized and positioned single blocks. Additionally, Chen \textit{et al.} introduced GridMask \cite{gridmask} to avoid erasing continuous areas, thus improving the localization ability of the resulting trained models.


However, these data augmentation methods differ fundamentally from the proposed decoupling operation in two aspects. Firstly, information deletion in object localization aims to encourage the network to \textit{focus more on the full image context} for identifying the objects, while the proposed DePatch targets to \textit{strengthen local patterns} of the generated patch. Secondly, these previous data augmentation approaches struggled with various deleting strategies to achieve a balance between excessive deletion and continuous region preservation \cite{gridmask}. In contrast, we introduce the border shifting and the progressive decoupling strategy to decouple all segments within the patch as much as possible, and thus encouraging them to be more robust against various degradations in complex real-world applications. Experimentally, we validate the superiority of our method in Section~\ref{sec:exp-aug}.

\begin{figure*}[ht]
	\centering
	\includegraphics[width=\linewidth]{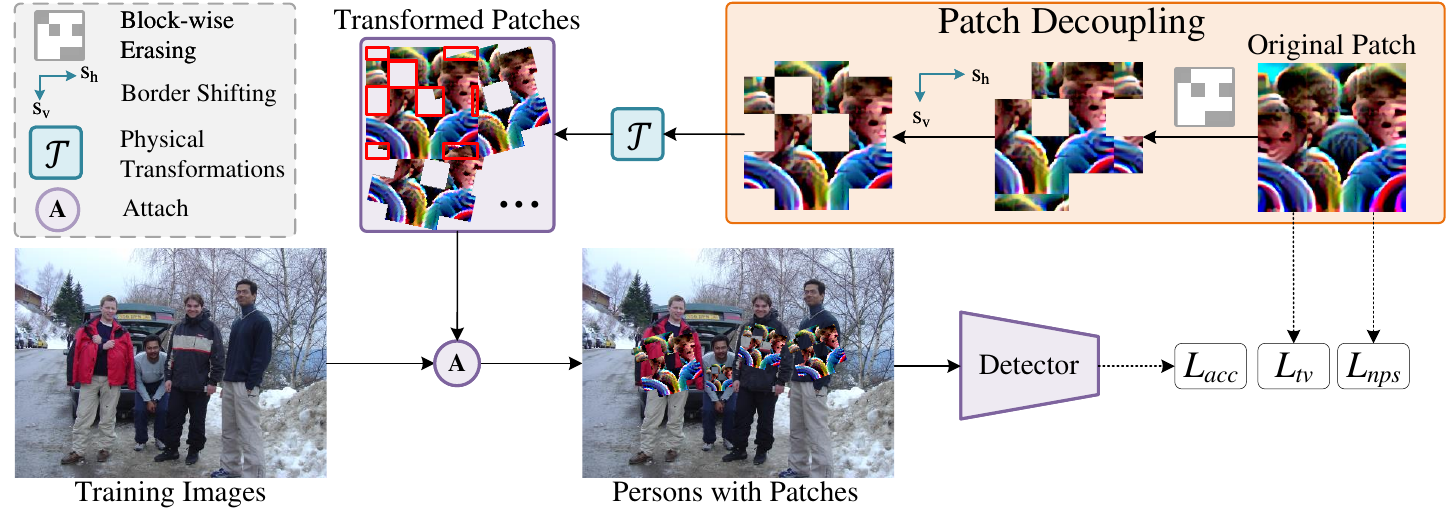}
	\caption{The overall pipeline for crafting the proposed DePatch. }
	\label{fig:pipeline}
\end{figure*}

\section{Overall Pipeline of Patch-based Attack}
As shown in Figure~\ref{fig:pipeline}, to achieve the goal of attacking person detectors in the real world, an adversarial patch is first processed by physical transformations. Then, it is optimized with the objective of lowering the accuracy of a specific person detector. 

\textbf{Patch Generation and Transformation.}
The adversarial patch $\mathbf{P}$ is initialized using Gaussian noise with a fixed shape and size. During iterative optimization, the pixel values of $\mathbf{P}$ are optimized via backpropagation.
For real-world applications, the physical transformations are deployed to $\mathbf{P}$ during optimization for handling complex physical conditions. Transformations such as rotation, contrast adjustment, brightness adjustment, and noise addition are incorporated and termed as Expectation over Transformation (EoT) \cite{advPatch}. To simulate clothing deformations through affine transformations, Thin Plate Spline (TPS) deformation \cite{advShirt} is also integrated into the simulated physical transformations. Toroidal Cropping (TC) technique \cite{advTexture} can be utilized as another physical transformation for clothing-based applications. Namely, employing TC ensures that the generated patch has up-down and left-right continuity, thus making it expandable and can cover clothing of any size. Here, we denote the operation of physical transformations as $\mathcal{T}(\cdot)$ and the transformed patch as $\tilde{\mathbf{P}} = \mathcal{T}(\mathbf{P})$.

\textbf{Attack Objectives.}
The transformed patch $\tilde{\mathbf{P}}$ is attached to the training images according to the labeled bounding boxes of the \textit{Person} class. Then, the attached images are detected by the frozen pre-trained detector. 
As the objective is to disable the detector in detecting persons with attached adversarial patches, existing methods consider reducing the object score of correct detections and then updating the pixels of the patch iteratively via backpropagation. Furthermore, the Non Printability Score (NPS) loss $L_{nps}$ and Total Variation (TV) loss $L_{tv}$ \cite{nps} are incorporated into the objective function to encourage the generated patch to be printable and smoother.

\section{Proposed Method}

\subsection{Patch Self-coupling}

\begin{figure}[t]
    \centering
    \includegraphics[width=0.6\linewidth]{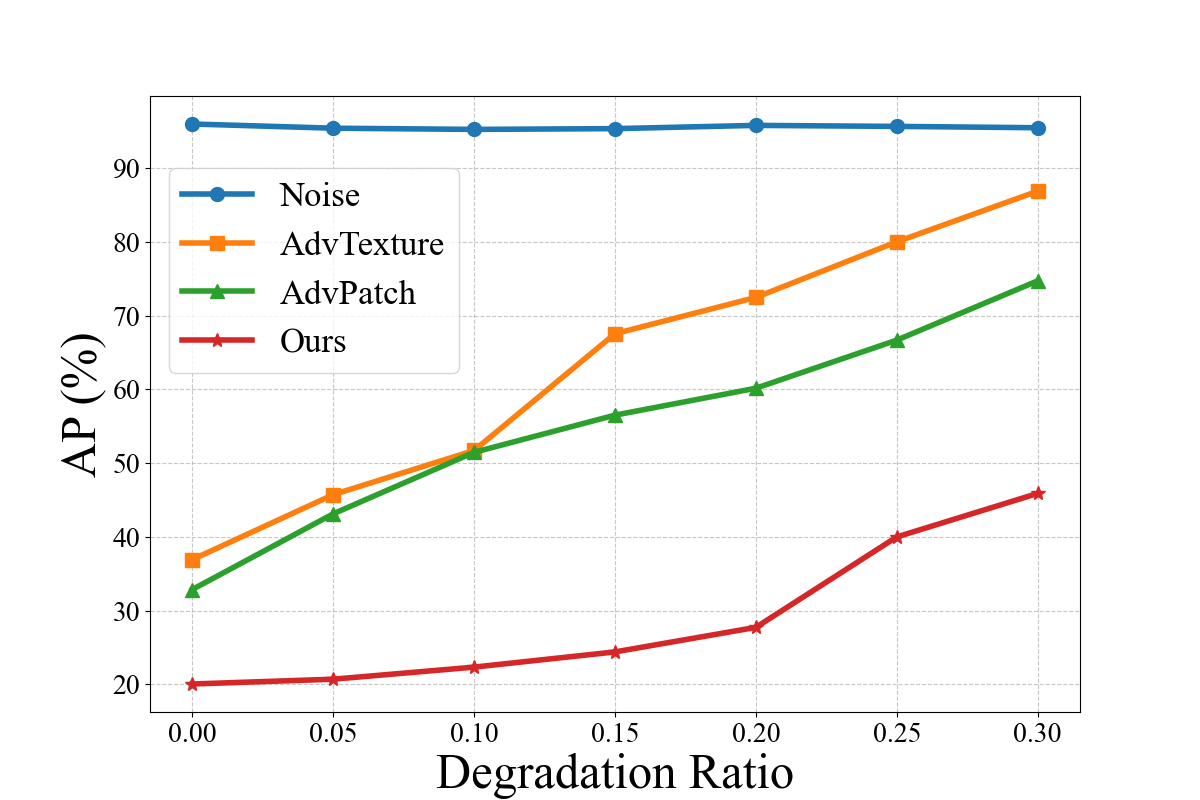}
    \caption{AP change of different patch-attack approaches on the Inria Person dataset under increasing degradation ratios}
    \label{fig:self-couple}
\end{figure}

To attack person detectors in real-world scenarios, the adversarial patches must be robust against physical transformations.
It is consistent with the expectations that some performance degradation will occur due to the inevitable information loss induced by the physical transformations. However, despite existing methods introducing various simulated physical transformations into the training process to improve robustness, they suffer from a sharp decline in performance even when encountering a slight degree of information loss, as shown in Figure~\ref{fig:self-couple}.

This abnormal decline indicates serious robustness issues with current methods, which we attribute to patch self-coupling. Specifically, since the entire patch is optimized simultaneously in every training step, its overall effectiveness is intrinsically dependent on every individual segment. That is, all segments are coupled with each other in exhibiting adversarial effectiveness. In other words, the patch is required to be intact for making a successful attack. 
Intuitively, self-coupling results in patches struggling to handle partial defects such as occlusions or contaminations. From a more considerable perspective, patches may be exposed to \textbf{non-uniform degradations} under diverse and complex physical conditions, such as noise, illumination changes, and pose changes. Given these challenging physical degradations may not entirely obliterate the adversarial pattern, the patches are expected to maintain effectiveness to a certain degree. 
However, the strict necessity for patch integrity inhibits the undamaged portions of the pattern from demonstrating adversarial effectiveness independently, thereby significantly impairing the attack robustness of existing methods in the real world.

\subsection{Decoupled Adversarial Patch}
\textbf{Block-wise Decoupling.} In order to improve the robustness of the physical adversarial patch in real-world scenarios, we propose the Decoupled adversarial Patch (DePatch) to address the self-coupling issue of the adversarial patch. 
Specifically, during each iteration of patch optimization, we stochastically set a ratio of the pixels to zero to reduce the inter-dependencies among various segments within the patch. 

Instead of performing stochastic decoupling among all pixels, which we term pixel-wise decoupling, we adopt block-wise decoupling by dividing the image into block-wise segments and stochastically erasing a certain ratio of segments. The reasons are two-fold:
1) Pixel-wise decoupling disrupts the relationship between each pixel and its surrounding pixels. This effect will drastically impact the spatial correlation of the generated patch, and it also causes the total variation loss (TV loss) to be unusable, which is often applied to encourage the generated patch to be smoother. Using block-wise decoupling, on the other hand, can maintain spatial correlation in the generated patch, except for a few minor partitions where segment border locations. Therefore, using block-wise decoupling ensures that the generated patches remain spatially correlated and can be smoothed by the TV loss.
2) Block-wise decoupling has the nature of simulating real-world occlusions and contamination, where a certain portion of continuous regions are often wrecked by hands, bags, or other objects in the environment. Hence, by introducing block-wise decoupling into the patch optimization process, the generated patch can better adapt to occlusion problems. Moreover, as each segment of the decoupled patch independently attains a certain level of attacking effectiveness, the intact patch can consequently achieve robust adversarial attack even when physical transformations severely disrupt a portion of the patch.

To apply block-wise decoupling, we first divide the adversarial patch $\mathbf{P}$ into $n \times n$ blocks, where $n$ can be considered as the granularity of the division. Hence, the set of divided blocks can be written as:
\begin{equation}
	\mathbf{P}=\{\mathbf{p}_{1},...,\mathbf{p}_{n\times n}\}, \label{eq_patch1}
\end{equation}
where $\mathbf{p}_{i}$ denotes the $i$-th divided block. Then, we introduce a binary mask $\mathbf{M}=\{\mathbf{m}_1,...,\mathbf{m}_{n\times n} \}$ where each element $\mathbf{m}_i$ is independently sampled from a Bernoulli distribution with a decoupling ratio $r$ of being set to zero.  Finally, we conduct element-wise multiplication:
\begin{equation}
	\hat{\mathbf{P}}=\mathbf{P} \odot \mathbf{M}. \label{eq_patch2}
\end{equation}
The adversarial patch with block-wise decoupling can be obtained by assembling $\hat{\mathbf{P}}$. Such operation of applying block-wise decoupling to patch $\mathbf{P}$ is denoted as $\mathcal{D}(\mathbf{P},n,r)$.


\textbf{Border Shifting.} Based on Eq.~\eqref{eq_patch1}, we successfully decoupled block-wise segments with each other in the same patch. However, given that the segmentation is predetermined based on granularity, pixels within the same segment are still simultaneously optimized and thus remain susceptible to the self-coupling issue.

Here, we further introduce border shifting inspired by the sliding window technique, which is widely employed in data streams \cite{slid1}, to ensure the decoupling of every possible segment. Specifically, we propose to shift the segment border of $\mathbf{M}$ randomly on the patch, rather than being confined to the fixed patch segments as described in Eq.~\eqref{eq_patch2}. However, directly shifting $\mathbf{M}$ leads to two issues: Firstly, due to the fragmentation and varying sizes of the segments after shifting $\mathbf{M}$, the calculation of applying fragmented masking segments via element-wise multiplication becomes complicated. Secondly, a portion of $\mathbf{M}$ will shift beyond the boundaries of the patch, thereby leading to an inaccurate representation of the decoupling ratio $r$.

For the first issue, we conduct border shifting on the assembled $\mathbf{M}$, and then we directly multiply the shifted $\mathbf{M}$ to the original $\mathbf{P}$. For the second issue, we concatenate the horizontal and vertical edges of $\mathbf{M}$ while shifting, which ensures the masked area that shifts out of the patch border can re-enter into $\mathbf{M}$ and thus maintains the accurate representation of $r$.
Such border shifting can be formalized as $\mathcal{S}(\mathbf{P},s_h,s_v)$, where $s_h$ and $s_v$ denote the horizontal and vertical distances of shifting, respectively. Hence, the decoupling operation $\mathcal{D}(\cdot)$ with border shifting can be written as:
\begin{equation}
	\mathcal{D}(\mathbf{P})=\mathcal{S}(\mathcal{D}(\mathbf{P},n,r),s_h,s_v).
\end{equation}
The final transformed patch attached to the persons during training is then formulated as:
\begin{equation}
	\Tilde{\mathbf{P}}=\mathcal{T}(\mathcal{S}(\mathcal{D}(\mathbf{P},n,r),s_h,s_v)).
\end{equation}
Notably, the erased segments in $\Tilde{\mathbf{P}}$ are considered transparent when $\Tilde{\mathbf{P}}$ is attached to the training images during training. That is, we preserve the image region under the erased segments instead of setting them as black. By doing this, we eliminate the possible adverse influence exerted by the black segments on the functioning adversarial pattern during training.


\begin{figure}[tbp]
		\centering
		\includegraphics[width=0.8\textwidth]{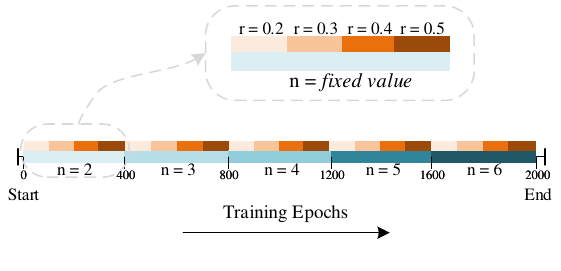}
		\caption{Illustration of the proposed progressive decoupling strategy (PDS) for improving attacking performance.}
		\label{fig:PDS}
\end{figure}
\textbf{Progressive Decoupling Strategy.} Block-wise decoupling has two variants: $n$ represents the granularity of block division, and $r$ denotes the decoupling ratio. An appropriate strategy for setting $r$ and $n$ can effectively enhance the attacking performance of DePatch. 
In this paper, we design a Progressive Decoupling Strategy (PDS) as illustrated in Figure~\ref{fig:PDS}, where $r$ and $n$ are adjusted progressively during different training stages. Considering that occlusions in the real world can significantly vary in size and position, training the patch to accommodate this variability by incorporating different values of $n$ into the optimization process could be beneficial. Meanwhile, to maintain relative stability within short training periods, the value of $n$ needs to be adjusted in a progressive step-by-step manner. To make the initial phase of training easier, we start with $n=2$ and gradually increase the granularity of the division to $n=6$.
As for $r$, we also adopt a scheme that gradually increases from $r=0.2$ to $r=0.5$ for an easier initialization of training. The adjustments for $r$ are implemented in cycles, where for each fixed value of $n$, we increase $r$ from 0.2 to 0.5.

\subsection{The Attack Objective Function}
To further enhance adversarial effectiveness, we propose a new optimization objective that better represents the accuracy of the predicted bounding boxes.
Specifically, the accuracy of a bounding box depends on its confidence score and Intersection over Union (IoU) with ground truth. 
Previous works \cite{advPatch, advShirt, advTexture} adopt the object score, which is comprised of the predicted class confidence and the predicted IoU, as the optimization target. In this paper, we argue that the real IoU could be a better target for optimization than the predicted IoU, and hence design an accuracy score for each bounding box.  Given a set of predicted bounding boxes $\{b_1, b_2,...,b_K\}$, their best IoUs and object scores can be written as $\{{IoU}_1, {IoU}_2,..., {IoU}_K\}$ and $\{{obj}_1, {obj}_2,..., {obj}_K\}$. Then we calculate $s$ as follows:
\begin{equation}
	s_k = w \cdot {IoU}_k + obj_k, \quad k = 1, 2, ..., K ,
\end{equation}
where $w$ is a trade-off parameter. Finally, we select the object score  corresponding to the highest accuracy score from the predicted bounding boxes, and it is then used as a part of the objective function to be minimized, denoted as $L_{acc}$:
\begin{equation}
	L_{acc} = {obj}_{\underset{k \in {1, 2, ..., K}}{ \operatorname{argmax}}(s_k)}    .
\end{equation}

Together with $L_{tv}$ and $L_{nps}$, we craft the proposed DePatch by minimizing the following attack objective function :
\begin{equation}
	DePatch = \underset{p}{\operatorname{argmin}}(L_{acc}+\alpha L_{nps}+ \beta L_{tv}) ,
\end{equation}
where $\alpha$ and $\beta$ are trade-off parameters. 
\begin{table*}[ht]
	\centering
 \caption{The APs (\%) on the Inria Person test set with simulating various physical transformations. \textcolor{red}{Red} and \textcolor{blue}{blue} indicate the best and the second-best performance, respectively.}
 \footnotesize
\begin{tabular}{lcccccccc}

\toprule
    Method      & Year       &   Original   & EoT & TC Mean  &Oc (0.1)&Oc (0.2) &Oc (0.3)& Overall \\ \midrule[0.7pt]
    Noise      &-&     96.23 & 96.09 & 95.71 & 96.38 &95.91& 95.52 & 96.12 \\
    AdvPatch \cite{advPatch} & 2019 &    \textcolor{blue}{18.55} & 32.87 & 81.89 & 51.47& 60.16& 74.72& 60.42\\
    AdvTshirt \cite{advShirt} & 2020 &     54.75 & 75.04 & 85.10 & 84.30&85.93& 90.75& 84.22\\
    AdvCloak \cite{advShirt2}  & 2020 &     57.11 & 76.63 & 82.62 & 82.15&87.45& 91.32& 84.04 \\
    AdvTexture \cite{advTexture} & 2022 &     25.96 & 36.52 & \textcolor{blue}{36.93} & 51.72&72.47& 86.88& 56.90\\ 
    T-SEA \cite{tsea} & 2023 &     20.21 & 27.18 & 78.78 & 36.31&48.43& 70.36& 46.71\\ 
        \midrule[0.7pt]
    DePatch & 2024 &    \textcolor{red}{17.75} & \textcolor{red}{20.06} & 54.45 & \textcolor{red}{22.35}&\textcolor{red}{27.74}& \textcolor{red}{45.90}& \textcolor{blue}{34.10}\\
    DePatch+TC  &  2024 &     21.43 & \textcolor{blue}{24.64} & \textcolor{red}{25.30} & \textcolor{blue}{28.12}&\textcolor{blue}{40.87}& \textcolor{blue}{49.26}& \textcolor{red}{33.78}\\
    \bottomrule
\end{tabular}

	\label{tab:main-digital}
\end{table*}
\begin{figure}[htbp]
	\centering
	\hspace{-3em}\includegraphics[width=0.8\textwidth]{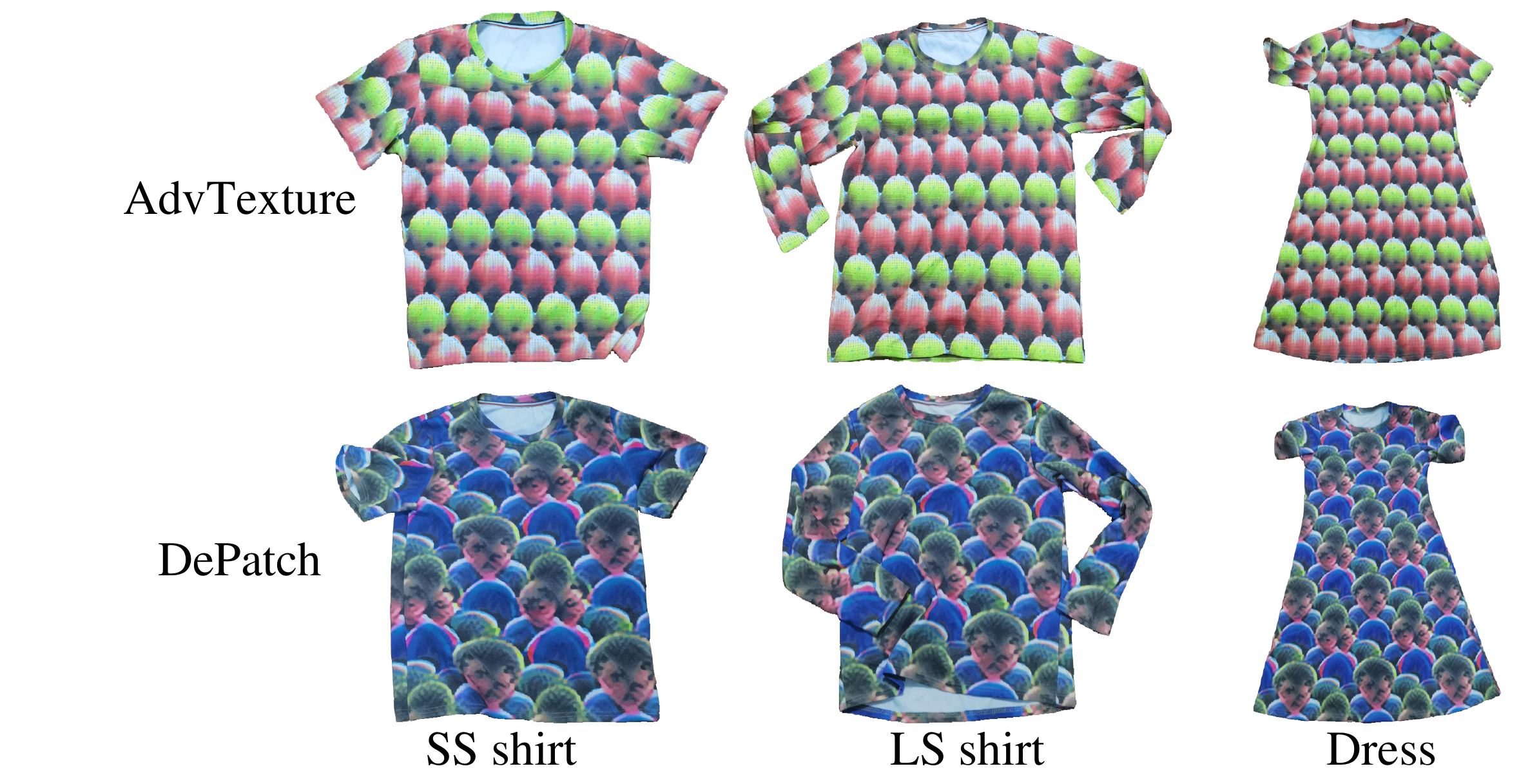}
	\caption{Three types of adversarial clothes crafted by covering them with expandable patches, \textit{i.e.}, AdvTexture \cite{advTexture} and DePatch. SS shirt, LS shirt, and Dress denote short-sleeve shirt, long-sleeve shirt, and dress used for evaluations, respectively. The colors are inevitably influenced by the lightness, camera ISO, and more physical conditions. }
	\label{fig:clothes}
\end{figure}
\begin{figure}[ht]
	\centering
	\begin{minipage}[t]{1\textwidth}
			\centering
			\begin{subfigure}[t]{0.155\textwidth}
				\centering
				\includegraphics[width=\textwidth]{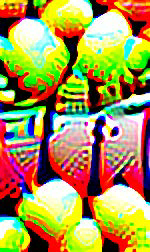}
				\caption{AdvTshirt \cite{advShirt}}
			\end{subfigure}
			\hfill
			\begin{subfigure}[t]{0.26\textwidth}
				\centering
				\includegraphics[width=\textwidth]{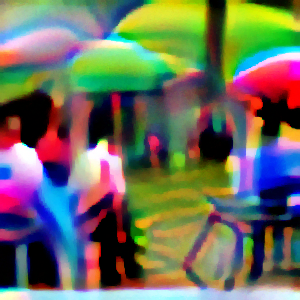}
				\caption{AdvPatch \cite{advPatch}}
			\end{subfigure}
			\hfill
			\begin{subfigure}[t]{0.26\textwidth}
				\centering
				\includegraphics[width=\textwidth]{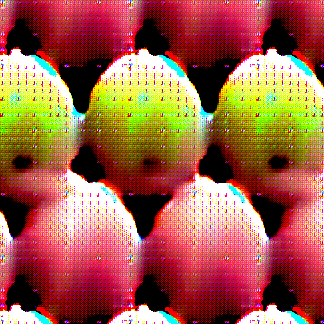}
				\caption{AdvTexture \cite{advTexture}}
				\label{fig:advtexture}
			\end{subfigure}
			\hfill
			\begin{subfigure}[t]{0.26\textwidth}
				\centering
				\includegraphics[width=\textwidth]{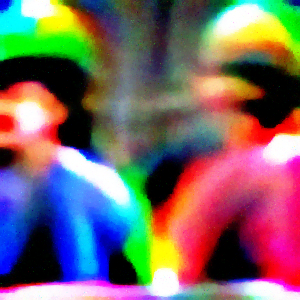}
				\caption{T-SEA~\cite{tsea}}
				\label{fig:advtexture-expanded}
			\end{subfigure}
			\begin{subfigure}[t]{0.155\textwidth}
				\centering
				\includegraphics[width=\textwidth]{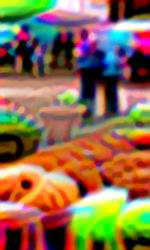}
				\caption{AdvCloak \cite{advShirt2}}
			\end{subfigure}
			\hfill
			\begin{subfigure}[t]{0.26\textwidth}
				\centering
				\includegraphics[width=\textwidth]{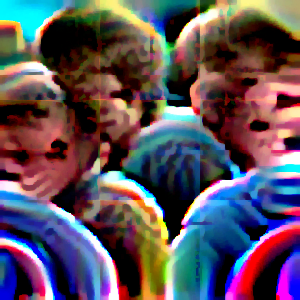}
				\caption{DePatch}
			\end{subfigure}
			\hfill
			\begin{subfigure}[t]{0.26\textwidth}
				\centering
				\includegraphics[width=\textwidth]{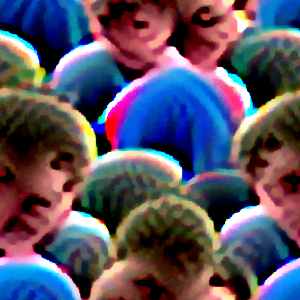}
				\caption{DePatch+TC}
				\label{fig:DePatch}
			\end{subfigure}
			\hfill
			\begin{subfigure}[t]{0.26\textwidth}
				\centering
				\includegraphics[width=\textwidth]{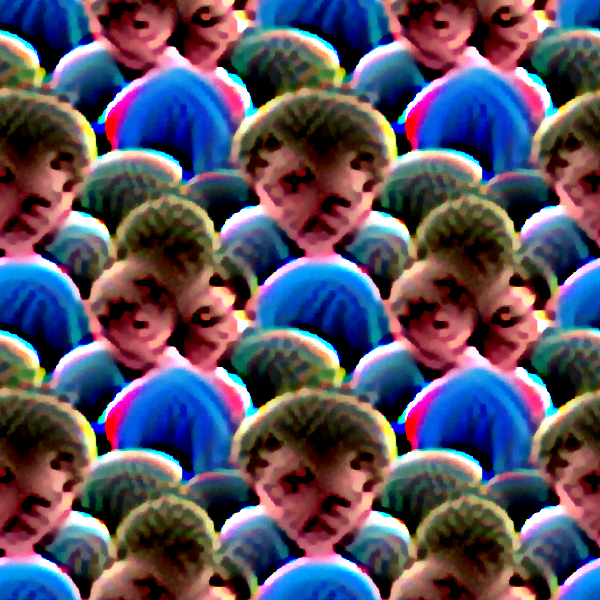}
				\caption{DePatch+TC (expanded)}
				\label{fig:DePatch-expanded}
			\end{subfigure}
			\caption{Adversarial patches produced by attacking YOLOv2. Figure~\ref{fig:DePatch-expanded} are expanded Figure~\ref{fig:DePatch} to craft clothes. For better layout presentation, both AdvTshirt and AdvCloak were proportionally scaled to maintain equal height in all patches. In fact, the actual patch area utilized during the evaluation was consistent across all patches for fair comparisons.}
			\label{fig:patches-v2}
	\end{minipage}
\end{figure}
\section{Experiments}
\subsection{Experimental Setup}

\textbf{Datasets.} Following \cite{advPatch, advTexture}, we train the proposed DePatch in the digital world on the Inria Person dataset \cite{inria}, which is a pedestrian dataset containing 614 images for training and 288 for testing. For cross-dataset evaluation, we collect images with the \textit{Person} class from the MS COCO validation set \cite{coco} and obtain 2693 images.
As for the physical world evaluation, we produce adversarial clothes (see Figure~\ref{fig:clothes}) and collect a real-world dataset. Specifically, we recruit three subjects (varying in age and height) to wear different adversarial posters or clothes in various complex physical conditions. Then, we record videos and extract 36 frames for each video. Therefore, each condition has 36 (frame) $\times$ 3 (subject) = 108 images for real-world evaluation. 

\textbf{Methods for Comparison.} We compare the proposed DePatch with AdvPatch \cite{advPatch}, AdvTshirt \cite{advShirt}, AdvCloak \cite{advShirt2}, AdvTexture \cite{advTexture}, and T-SEA~\cite{tsea}. We additionally provide a version of DePatch deploying the TC technique during optimization, which is recursive and can be applied for clothing-based applications, denoted as DePatch+TC. We obtain the baseline adversarial patches by using their official codes \cite{advPatch,advTexture,tsea} or directly extracting them from their papers \cite{advShirt,advShirt2}. The patterns of each compared patch are shown in Figure~\ref{fig:patches-v2}. DePatch is initialized with a $300\times300$ shape and Gaussian noise. We employ the Adam optimizer \cite{adam} for patch optimization and conducted training for 2000 epochs with a learning rate of 0.03. The trade-off parameters are set as $w=3$, $\alpha=0.01$, and $\beta=2.5$. All experiments are implemented in PyTorch on one NVIDIA Tesla V100 GPU and all patches are resized to encompass 90000 pixels for fair comparison.

\textbf{Evaluation Metrics.} In the digital world, we adopt average precision (AP) as the evaluation metric and set the Intersection over Union (IoU) threshold at 0.5. AP can represent the precisions and confidences of the detectors, and a smaller AP indicates a stronger adversarial attack. The bounding boxes predicted on the clean images with confidences higher than 0.5 are treated as the ground truth bounding boxes, that is, the AP of detecting clean images is set to approximately 1.0. In the physical world, the attack success rate (ASR) is used as the evaluation metric. ASR denotes the percentage of frames within a video where the detector yields an incorrect detection, and a higher ASR suggests a better performance of the physical attack. We consider a frame as incorrectly detected if it has a confidence score lower than 0.5 or an IoU lower than 0.5. 

\textbf{Object Detectors.} Since the baseline methods were mainly trained and tested on attacking YOLOv2 \cite{yolov2}, we also select YOLOv2 as our primary test bed to ensure a fair comparison. To further verify the robustness and effectiveness of our method, we conduct the transfer study within the one-stage detectors, that is, YOLOv2, YOLOv3 \cite{yolov3}, and YOLOv5 \cite{yolov5}.

\begin{table}[b]
\centering
	\caption{The APs (\%) on the MS-COCO validation set. Oc Mean denotes the mean result of occlusions with ratios 0.1 to 0.3.}
\small
\begin{tabular}{lcccc}
		\toprule
Method       & Original & EoT & Oc Mean&Overall \\		\toprule
Noise        & 89.18   & 90.91 & 87.17&89.04\\
AdvPatch \cite{advPatch}    & 38.15   & 53.26  & 54.33&53.80 \\
AdvTshirt \cite{advShirt}  & 53.98   & 55.33 & 68.48 &61.91 \\
AdvCloak \cite{advShirt2}  & 64.73   & 65.17 & 72.23 &68.70\\
AdvTexture \cite{advTexture}  & 38.01   & 51.09 & 59.94&55.52 \\
T-SEA \cite{tsea}  & 47.31   & 66.59 & 74.54&7.57 \\
DePatch      & \textbf{30.37}   & \textbf{32.16}  &  \textbf{38.02} &\textbf{35.09}\\ \bottomrule
\end{tabular}

\label{tab:coco}
\end{table}

\begin{table*}[ht]
\centering
\caption{The ASRs for the\textbf{ poster-based }applications in the real world. Notably, T-SEA is not designed for real-world applications.}
\footnotesize
\setlength{\tabcolsep}{3pt} 
\begin{tabular}{lccccccccc}
	\toprule
	\multirow{2}{*}{Method}&\multicolumn{3}{c}{Distances}&\multicolumn{4}{c}{Occlusions}&\multirow{2}{*}{Outdoor}  &\multirow{2}{*}{Overall}\\ 
	\cmidrule(r){2-4}\cmidrule(lr){5-8}
	& 1.5m&3.0m&4.5m& Oc (0.1) & Oc (0.3) & SH & BH &  &\\ \toprule
AdvPatch \cite{advPatch}    &0.1204&0.5926&0.0000& 0.0185  & 0.0000  & 0.0208& 0.0000 &0.2500   & 0.1253\\ 
AdvTshirt \cite{advShirt}    &0.2222&0.6667&0.1667& 0.3611  & 0.1687  & 0.6875 & 0.0833 &0.0833   & 0.3177\\ 
AdvTexture \cite{advTexture}  &0.5648&0.8148&0.1667& 0.5556  & 0.2500  & 0.7037& 0.2917&0.3611  & 0.4694\\ 
T-SEA \cite{tsea}  &0.2500&0.4629&0.0000& 0.3425  & 0.1667  & 0.7500& 0.2917&0.1296  & 0.2867\\ 
DePatch     &\textbf{0.7037}&\textbf{0.8981}&\textbf{0.8333}& \textbf{0.8889}  & \textbf{0.7500 } &\textbf{ 0.9583 }     & \textbf{0.8125}&\textbf{0.5677}  & \textbf{0.8014}\\ 
\bottomrule
\end{tabular}

   \label{tab:occlusion-physical-poster}
\end{table*}

\begin{table*}[ht]

\centering
			\caption{The ASRs for the \textbf{clothing-based} applications in the real world.}
\footnotesize
\begin{tabular}{lcccccccc}

	\toprule
	\multirow{2}{*}{Method}&\multicolumn{3}{c}{Distances}&\multicolumn{2}{c}{Occlusions}&\multirow{2}{*}{$360^{\circ}$}&\multirow{2}{*}{Outdoor}  &\multirow{2}{*}{Overall}\\ 
	\cmidrule(r){2-4}\cmidrule(lr){5-6}& 1.5m&3.0m&4.5m& SH & BH &&  &\\\midrule
                                 AdvTexture \cite{advTexture}& 0.7778&1.0000&0.3333&0.7708&	0.5463&0.8055&0.6296   &0.6947\\  
                             DePatch& 1.0000&1.0000&0.8642&1.0000 &0.8550&0.9629&0.6852    &0.9096\\\bottomrule
\end{tabular}

			\label{tab:occlusion-physical-clothing}
\end{table*}
\subsection{Results in the Digital World}

\subsubsection{Results on the Inria Person dataset.}
As shown in Table~\ref{tab:main-digital}, we evaluate the adversarial effectiveness of different patch-based methods in various attack settings of simulated physical transformations on the Inira Person test set. \textit{Original} denotes using patches without any transformations.
\textit{EoT} denotes the patches are transformed by EoT (\textit{i.e.}, noise, contrast, brightness, and rotation) \cite{advPatch}.
Notably, all experiments conducted in the digital world except \textit{Original} are additionally transformed by EoT.
\textit{TC Mean} denotes that the patches are randomly resampled via the Toroidal Cropping technique \cite{advTexture} during evaluation, indicating patch expandability. Due to the randomness introduced by the TC resampling,  we performed $10$ repeated experiments for each attack method and reported the mean values in the result intervals. \textit{Oc (r)} denotes simulating different levels of real-world occlusions via cutout \cite{cutout} by masking ratio $r$ of the patches during evaluation. We deploy random masking following \cite{cutout} due to its nature of resembling occlusions. \textit{Overall} reports the mean values across different settings except \textit{Original}, and hence it can be treated as the overall performances against different simulated physical transformations. \textit{Noise} denotes a patch filled with Gaussian noise. 

Due to the patch self-coupling issue, both AdvPatch \cite{advPatch} and AdvTexture \cite{advTexture} exhibit comparable adversarial effectiveness only when no transformations are applied. Their performance degrades dramatically when any simulated physical transformation is deployed, which indicates their inadequate robustness. 
Since AdvTshirt \cite{advShirt} is trained on a private dataset and AdvCloak \cite{advShirt2} is trained on a modified MS COCO dataset \cite{coco}, their performance on the Inria test set is poor. Still, abnormal drops in attacking performances are presented when facing simulated physical transformations.
In contrast, DePatch achieves the best adversarial effectiveness in non-expandable scenarios. Specifically, it demonstrates robustness when dealing with various physical transforms, that is, its decline in adversarial effectiveness is notably less significant. The promising performance in TC Mean demonstrates the expandability of DePatch+TC,  thus allowing it to be deployed for clothing-based applications. However, while DePatch+TC shows enhancements over AdvTexture, its performance under different conditions is outperformed by DePatch, which indicates that the TC technique does not provide positive influences on adversarial effectiveness. 

\subsubsection{Cross-dataset evaluations}

Following the same evaluation protocol on the Inira Person test set, we assess the patches trained on the Inria Person dataset using the MS COCO validation set \cite{coco} and the results are shown in Table~\ref{tab:coco}. Notably, deploying the Gaussian noise patch can drop APs to 0.89, which is attributed to the presence of numerous incomplete human figures in the data. This indicates that Inria Person may be more appropriate than MS COCO for the task of attacking person detectors. Nevertheless, DePatch can still maintain effectiveness when resisting EoT and different occlusions and consistently outperforms other methods in all conditions.

\subsection{Results in the Real World}

\begin{figure*}
	\centering
	\includegraphics[width=1\textwidth]{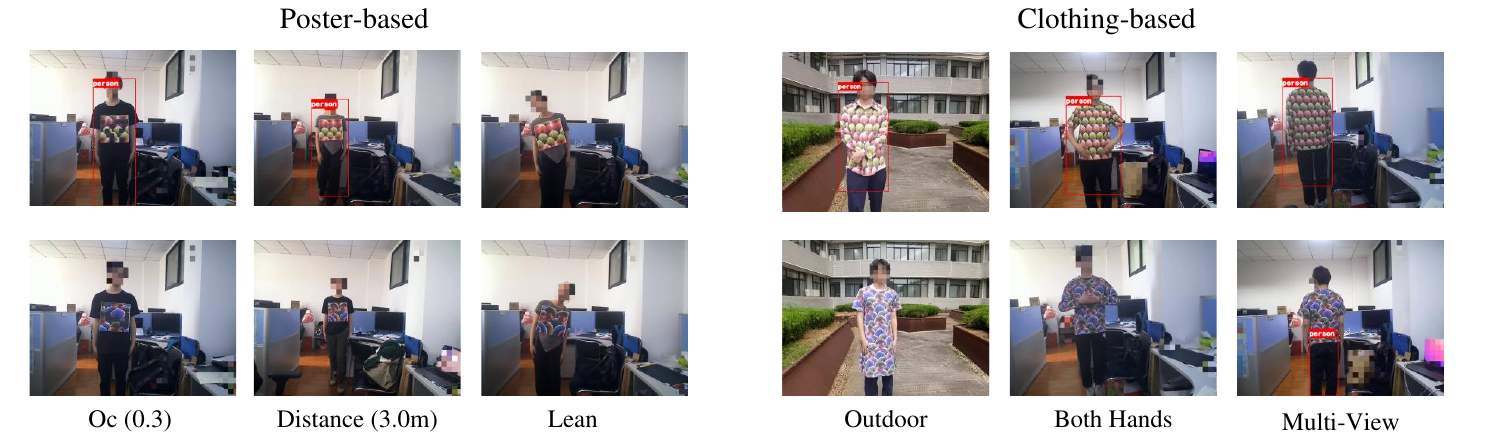}
	\caption{Examples of real-world data in the crafted dataset. The upper row is the patch generated by AdvTexture~\cite{advTexture}, and the lower row is ours. We have filtered out bounding boxes of non-person classes for a clearer illustration, and the mosaics are added after detection for privacy protection.}
	\label{fig:dataset-visual}
\end{figure*}

Here, we evaluate the adversarial effectiveness in real-world scenarios.
We introduce two different forms of implementing adversarial patches in the physical world: 1) Poster-based application: printing adversarial patches on posters and then attaching the posters to the front of T-shirts following the protocol of AdvPatch \cite{advPatch}. 2) Clothing-based application: fabricating clothes covered with expandable adversarial patches using digital textile printing following the protocol of AdvTexture \cite{advTexture}. 
The dataset contains people wearing different adversarial clothes and posters at distances of 1.5m, 3.0m, and 4.5m, with each distance featuring various postures (\textit{e.g.}, spin and lean). In addition, we perform adversarial attacks with more complex physical conditions including Occlusions, 360$^{\circ}$, and Outdoor. Specifically, \textit{SH} and \textit{BH} denote occluding the adversarial poster and clothing by a single hand and both hands \cite{advShirt,advShirt2}, and 360$^{\circ}$ denotes deploying adversarial clothing to attack the detectors from 360$^{\circ}$ viewing angles. 

Firstly, we provide image examples of our real-world dataset in Figure~\ref{fig:dataset-visual} for visual impressions. Due to the different body metrics of each subject, we minorly adjusted the relative positions of each subject, while we ensured that the positions and lighting were consistent for the same group of comparisons. It can be observed that previous methods cannot consistently maintain robustness in both forms of real-world applications. 
One limitation of our method is that it sometimes causes the detectors to generate the \textit{Person} class bounding boxes with incorrect positions, which is not ideal for the purpose of making a person invisible from person detectors.

\textbf{Poster-based Applications.}
As shown in Table~\ref{tab:occlusion-physical-poster}, all methods exhibit optimal adversarial performance at a distance of 3.0m. However, at a distance of 4.5m, the performance of all patches suffers a drastic decline, with AdvPatch and AdvTshirt diminishing so severely that they nearly lose their adversarial effectiveness. T-SEA is not designed for physical-world applications, hence it suffers from the physical transformations that severely undermine its attacking performance. Nevertheless, our approach exhibits relatively stable and promising adversarial effectiveness across the three evaluated distances.
For the more complex physical conditions, AdvPatch, AdvTshirt, and AdvTexture are significantly affected by occlusions, that is, they nearly lose their adversarial effectiveness with a slight occlusion ratio. In contrast, our methods are robust to slight occlusions (\textit{i.e.}, Oc (0.1) and SH) and can even maintain robustness under severe occlusions (\textit{i.e.}, Oc (0.3) and BH). Meanwhile, the proposed method can outperform other methods in the outdoor scenario despite suffering from stronger degradation compared with the indoor cases. 

\textbf{Clothing-based Applications.}
In Table~\ref{tab:occlusion-physical-clothing}, we present the mean results among three types of clothing-based applications, that is, dress, long-sleeve, and short-sleeve shirts. It can be observed that DePatch has outperformed AdvTexture across all evaluated physical conditions, which indicates that the proposed decoupling operations can be effectively deployed for clothing-based applications. Specifically, DePatch can achieve 100\% ASR at distances of 1.5m and 3.0m.  This is consistently observed in scenarios that include occlusions with a single hand (SH) and the $360^{\circ}$ multi-view. While in the conditions where all poster-based applications underperform (\textit{i.e.}, at 4.5m, with BH, and outdoor), the adversarial clothes crafted using DePatch can still exhibit superior adversarial effectiveness.

\begin{table*}[htbp]
    \centering
    \small
            \caption{Ablation study for PDS, border shifting (BS), and $L_{acc}$. The \ding{51} denotes the deployed component for each ablation variant.}
                \small
            \begin{tabular}{l@{\hspace{1em}}c@{\hspace{0.2em}}c@{\hspace{0.2em}}c@{\hspace{1em}}c@{\hspace{0.5em}}c@{\hspace{0.5em}}c@{\hspace{0.5em}}c@{\hspace{0.5em}}c@{\hspace{0.5em}}c}
                \toprule
                Method  & $L_{acc}$ & PDS & BS & EoT & Oc (0.1) & Oc (0.2) & Oc (0.3) & Overall\\
                \toprule
                Var-A & & &  & 31.05 & 40.59 & 47.66 & 56.58 & 43.97 \\
                Var-B & & \ding{51} & \ding{51} & 24.76 & 31.42 & 39.45 & 51.77 & 36.85 \\
                Var-C & \ding{51} & \ding{51} &  & 24.15 & 29.45 & 38.53 & 48.71 & 35.21 \\
                Var-D & \ding{51} & & \ding{51}  & 22.25 & 27.52 & 34.19 & 47.86 & 32.95 \\ 
                \midrule
                DePatch & \ding{51} & \ding{51} & \ding{51} & 20.06 & 22.35 & 27.74 & 45.90 & 29.01 \\
                \bottomrule
            \end{tabular}
            \label{tab:abl}
\end{table*}
\begin{table*}[htbp]
            \centering
            \caption{The APs (\%) of replacing PDS with simpler variant-adjusting strategies.}
                \small
            \begin{tabular}{lc@{\hspace{0.5em}}c@{\hspace{0.5em}}c@{\hspace{0.5em}}c@{\hspace{0.5em}}c}
                \toprule
                & EoT & Oc (0.1) & Oc (0.2) & Oc (0.3) & Overall\\
                \toprule
                S$_{n2}$ & 26.00 & 32.31 & 37.16 & 44.11 & 34.89\\
                S$_{n4}$ & 24.89 & 30.45 & 37.50 & 48.31 & 35.29\\
                S$_{n6}$ & 21.77 & 26.40 & 30.32 & 47.75 & 31.56\\
                \midrule
                PDS & 20.06 & 22.35 & 27.74 & 45.90 & 29.01\\
                \bottomrule
            \end{tabular}
            \label{tab:pds}
\end{table*}
\subsection{Ablation study}
In this section, we study the impact of the proposed PDS, border shifting (BS), and $L_{acc}$ on the Inria Person test set, and report APs (\%) under various circumstances to provide a more comprehensive understanding of the proposed DePatch.
In Table~\ref{tab:abl}, we design four more ablation variants of the proposed DePatch (\textit{i.e.,} Var-A to Var-D) for ablation study. It can be observed that $L_{acc}$ enhances the adversarial effectiveness under various attack settings, while PDS further mitigates the self-coupling issue of adversarial patches. Meanwhile, BS simulates more fragmented occlusion scenarios and ensures more extensive decoupling, thereby demonstrating its significance in occlusion conditions. 

Moreover, to verify the necessity of the proposed PDS, we introduce simpler strategies for adjusting the decoupling variants. Namely, we set the granularity $n$ to fixed values (\textit{i.e.}, $n=2,4,6$), and the ratio $r$ is randomly selected from 0.2 to 0.5. These strategies are denoted by S$_{n2}$, S$_{n4}$, and S$_{n6}$, respectively. As shown in Table~\ref{tab:pds}, S$_{n2}$ exhibits limited decoupling effectiveness due to its coarse divisions. Therefore, its performance in resisting EoT and small occlusions is poor while performs slightly better when occlusion is relatively large. S$_{n4}$ and S$_{n6}$ are finer-grain divisions and hence their decoupling effectiveness is improved. Still, they fail to tackle different ratios of occlusion consistently and their adversarial effectiveness is outperformed by PDS. 
\begin{figure*}[htbp]
	\centering

 	\begin{subfigure}[b]{0.32\textwidth}
		\includegraphics[width=\textwidth]{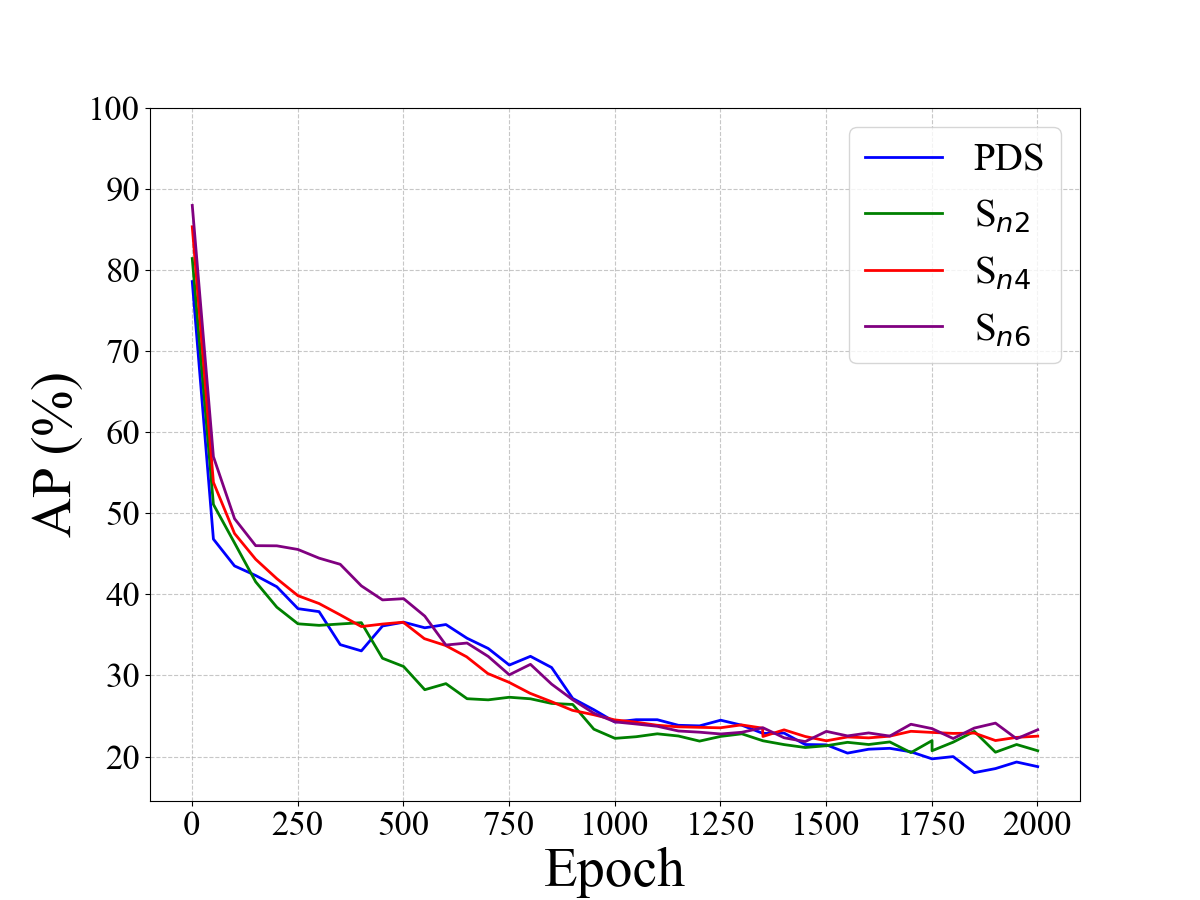}
		\caption{EoT}
		\label{fig:pdscurve_eot}
	\end{subfigure}
	\hfill
	\begin{subfigure}[b]{0.32\textwidth}
		\includegraphics[width=\textwidth]{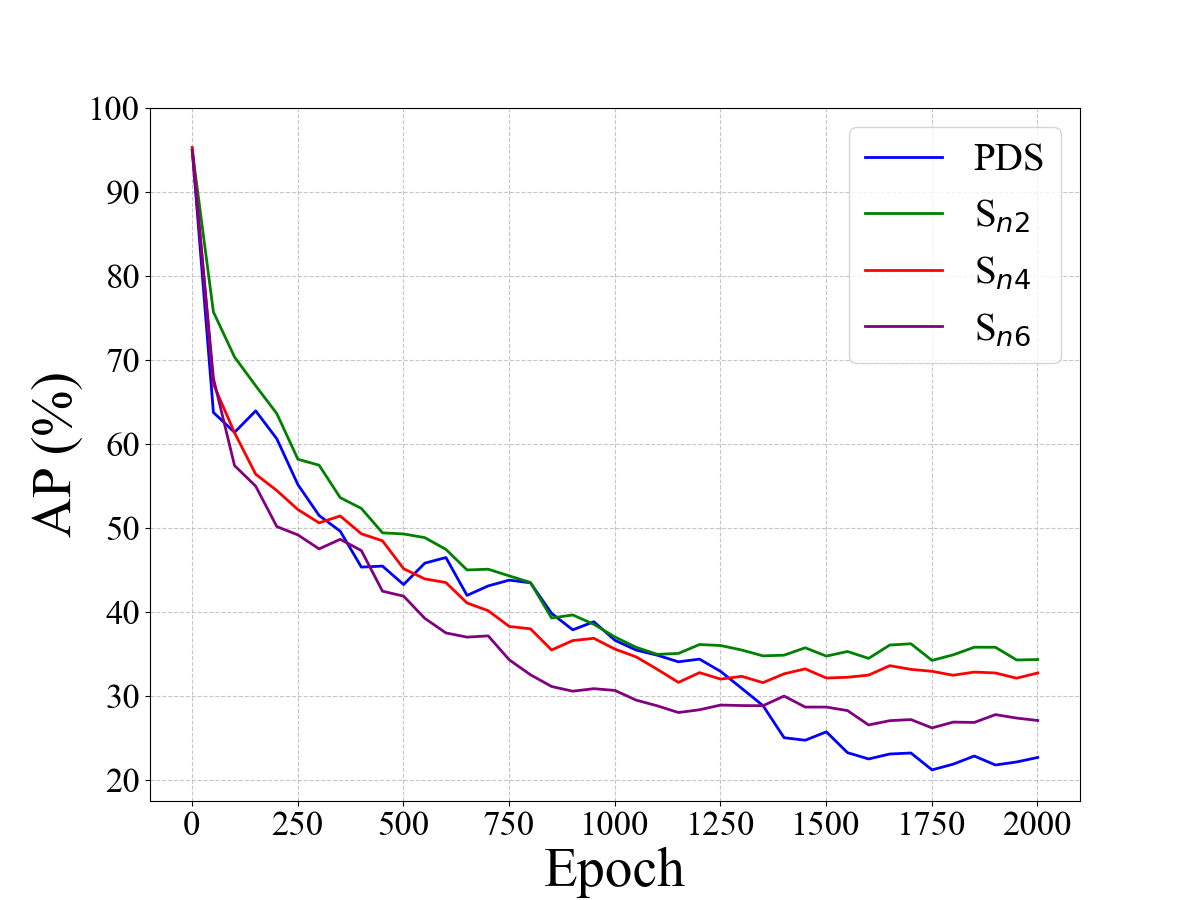}
		\caption{Oc (0.1)}
		\label{fig:pdscurve_oc1}
	\end{subfigure}
 	\hfill
	\begin{subfigure}[b]{0.32\textwidth}
		\includegraphics[width=\textwidth]{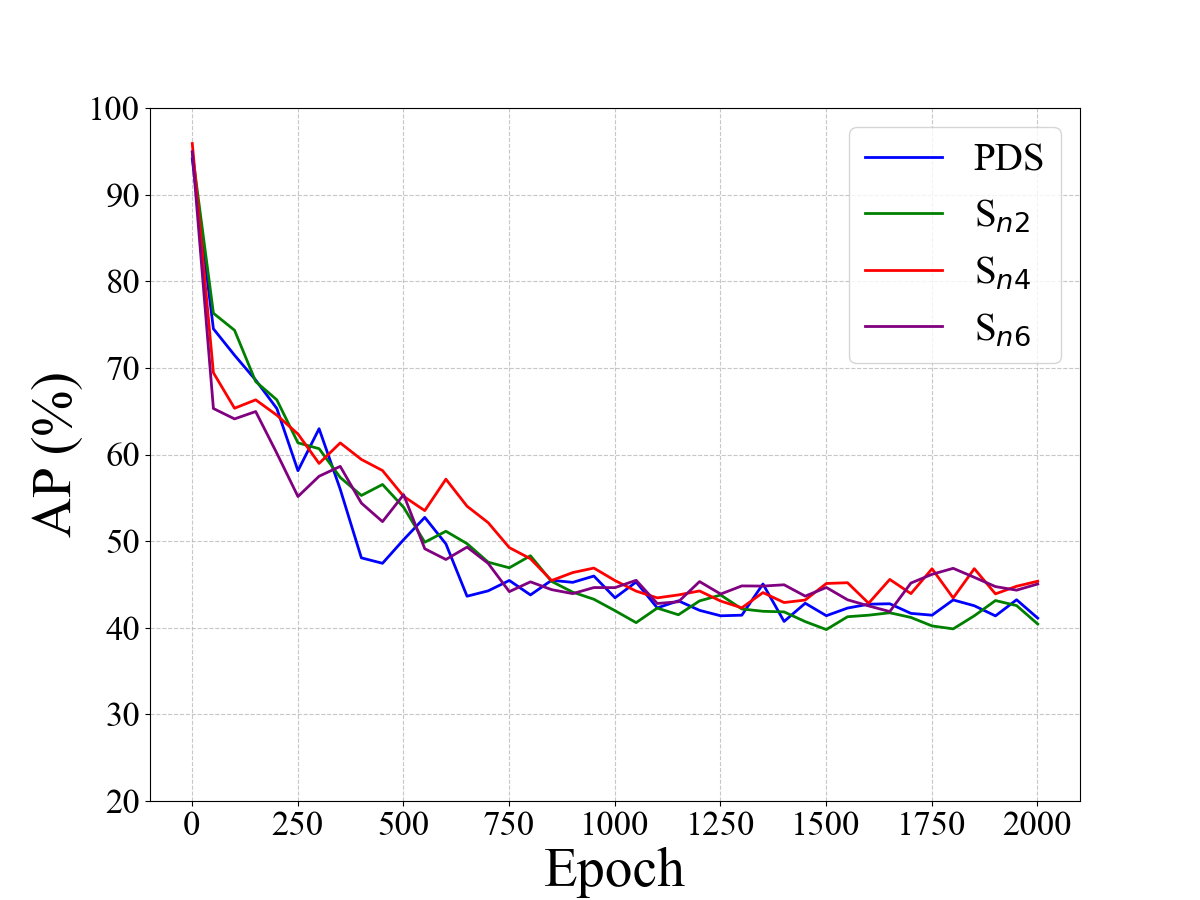}
		\caption{Oc (0.3)}
		\label{fig:pdscurve_oc3}
	\end{subfigure}
	\caption{Performance variation of patches with different strategies during training.}
    \label{fig:pdscurve}
\end{figure*}

\begin{table*}[ht]

	\centering
		
		\caption{The APs (\%) of different detectors for transferability evaluations in the digital world. \textit{Source} denotes the source detector used for training the adversarial patches. }
  \footnotesize
	\begin{tabular}{lccccccc}
		\toprule
		Method & Source & YOLOv2 &YOLOv3 & YOLOv5 & FR & MR & RetinaNet \\
		\midrule
        AdvTexture \cite{advTexture} & \multirow{3}[2]{*}[0.5ex]{YOLOv2} & 36.52 & 87.98 & 90.72 & 75.50 & 70.85 & 71.09 \\
        DePatch &  & 20.06 & 32.09 & 60.41 & 55.00 & 53.70 & 66.13 \\
        DePatch+TC &  & 24.64 & 27.94 & 52.65 & 39.99 & 51.47 & 45.19 \\
        \midrule
        DePatch & \multirow{2}[2]{*}[0.5ex]{YOLOv5} & 38.56 & 56.16 & 40.32 & 53.68 & 51.45 & 64.72 \\
        DePatch+TC &  & 35.35 & 39.85 & 35.76 & 39.11 & 47.72 & 40.44 \\
        \midrule
        DePatch & \multirow{2}[2]{*}[0.5ex]{FR} & 34.99 & 25.67 & 39.37 & 22.88 & 25.93 & 29.01 \\
        DePatch+TC &  & 43.92 & 50.53 & 50.27 & 28.76 & 35.34 & 36.68 \\
		\bottomrule
	\end{tabular}

	\label{tab:trans-digital}
\end{table*}

In addition, we provide the performance variation of patches on the training set when trained using PDS and \{S$_{n2}$,S$_{n4}$,S$_{n6}$\}. Specifically, starting from epoch 1, we store an intermediate result of a patch every 50 epochs, and then evaluate AP on the training set. The reason we employ AP rather than Accuracy Loss ($L_{acc}$) is that the loss is calculated under varying conditions of decoupling variants, undermining its reliability for horizontal comparisons. Conversely, the attack effectiveness of complete patches on the training set can be considered unbiased as a reference for the convergence status. As shown in Figure~\ref{fig:pdscurve_eot}, it can be observed that the strategies with smaller granularity demonstrate faster performance improvements. This is attributed to the inadequate decoupling of pixels in larger segments, which simplifies the learning process and consequently leads to rapid overfitting on the training set. However, when employing PDS, the inconsistent value of $n$ during the training process leads to performance fluctuation and a slower speed of improvement. Nevertheless, PDS ultimately reached a more favorable state of convergence compared to alternative strategies.
In Figure~\ref{fig:pdscurve_oc1}, the performance of PDS during the initial phase of training is suboptimal. Progressively adjusting the ratio from low to high yields a marginally improved outcome compared to S$_{n2}$, yet it remains inferior to the results of S$_{n4}$ and S$_{n6}$. Nevertheless, with the gradual increase in the granularity of PDS, it ultimately exhibits superior performance. In Figure~\ref{fig:pdscurve_oc3}, the increased occluded area contributes to greater uncertainty, resulting in more dramatic fluctuations in performance. S$_{n2}$, which employs a uniformly coarse granularity division similar to Oc (0.3), demonstrates the most effective performance. PDS also exhibits promising effectiveness by covering a wide range of granularities.

\begin{table}[htbp]
\centering
	\caption{The APs (\%) of replacing the proposed decoupling operation with information deletion methods.}
 \footnotesize
\begin{tabular}{lc@{\hspace{0.65em}}c@{\hspace{0.65em}}c@{\hspace{0.65em}}c@{\hspace{0.65em}}c}
		\toprule
Method&  EoT&Oc (0.1)& Oc (0.2)& Oc (0.3) & Overall\\\toprule
Vanilla \cite{advPatch}& 32.87& 51.47& 60.16& 74.72&54.81\\		 \midrule
HaS \cite{HaS}&  32.59&39.01& 46.35& 57.61& 43.89\\
Cutout \cite{cutout}&  39.31&47.41& 53.55& 61.07& 50.33\\
GridMask \cite{gridmask}&  37.88&51.81& 66.18& 78.28& 58.54\\
RE \cite{randomerasing}& 24.32& 33.01& 43.95& 52.50&38.45\\		 \midrule
Ours&  20.06& 22.35&27.74& 45.90& 29.01\\
\bottomrule
\end{tabular}

\label{tab:InforDel}
\end{table}
\subsection{Comparison with Directly Implementing Information Deletion}~\label{sec:exp-aug}
In Section~\ref{sec:aug}, we analyzed how our method is \textit{fundamentally distinct} from information deletion from a theoretical perspective. Here, we additionally present experimental results to validate the analysis. Following the official code in implementation, we deploy four existing information deletion methods \cite{HaS,cutout,gridmask,randomerasing} to replace the proposed decoupling operation. AdvPatch \cite{advPatch}, being trained without any masking, is termed as \textit{Vanilla} and is utilized as the baseline in the comparison. In Table~\ref{tab:InforDel}, the performance using these methods is outperformed by using decoupling. Moreover, the performance of GridMask is lower than Vanilla in all evaluation metrics. Similarly, Cutout and HaS perform comparably to Vanilla, indicating that these three methods cannot improve the effectiveness of adversarial patches. 
Due to the masks applied by RE exhibiting more random positions, shapes, and sizes,  there is a discernible enhancement in its effectiveness compared with the baseline. However, its poor performance in countering various occlusions demonstrates that it still suffers from the self-coupling issue. However, its poor performance in the face of various occlusions demonstrates that it still suffers from the self-coupling issue. Therefore, we substantiate the specificity of the decoupling method in enhancing patch robustness and its fundamental distinction from information deletion.



\subsection{Transfer Study} Since the detectors in real-world applications are usually unknown, transferability is an important factor for the physical adversarial examples to be applicable. 
We conduct training on YOLOv5 \cite{yolov5} and Faster R-CNN \cite{faster-rcnn} in addition to YOLOv2 \cite{yolov2} and perform evaluations on a much wider range of mainstream detectors including Faster R-CNN (FR), Mask R-CNN (MR) \cite{mask-rcnn}, and RetinaNet \cite{retinaNet}, respectively. Table~\ref{tab:trans-digital} presents the results of the transfer experiments. AdvTexture exhibits weak transferability, losing nearly all its adversarial effectiveness when applied to other detectors. In contrast, the proposed DePatch demonstrates outstanding transferability within the YOLO or R-CNN series respectively.  We consider that the effectiveness distinctions between the YOLO series and the R-CNN series may be due to their instinct architecture robustness. Specifically, the R-CNN series produce much more bounding boxes (i.e., object candidate boxes for selection) than the YOLO series \cite{detector_survey}. Therefore, more boxes indicate more chances to produce correct predictions, thereby harder to be evaded.  To enhance the effectiveness of crossing detectors with completely different architecture, improved optimization objectives or model ensemble techniques may be necessary. In addition, we provide a \textit{Supplementary Video} that showcases subjects wearing adversarial clothes in various casual and realistic postures.


\begin{table}[ht]
\centering
\caption{Performance on DePatch against the defensive methods (AP).}
\footnotesize
\begin{tabular}{ccccc}
    \toprule
    Defenses   & Original   & EoT &  Oc Mean & Overall\\
    \midrule
    w/o Defense    & 17.75  & 20.06 & 32.00 &  26.03  \\
    UDF     & 33.39 & 35.01 &  44.15 & 38.58\\
    JPEG & 27.16  & 27.90 & 35.45 & 31.68\\
    \bottomrule
\end{tabular}
\label{defense}
\end{table}
\subsection{Effectiveness Against Defensive Methods.}
The individual functionality of DePatch also enhance its effectiveness against adversarial defenses.
In Table~\ref{defense}, we conduct evaluations against adversarial defense methods that can be implemented to the physical-world person detectors, that is, JPEG compression and Universal Defensive Frame (UDF)~\cite{UDF}. JPEG is similar to the simulated physical transformations that are included during training and hence our method exhibits consistent performance. UDF is a very strong defense method that adds designed defensive patterns as the padding of input frames, attempting to destroy the adversarial pattern in the intact patches. Attributes to the decoupling operation that enables segments with individual effectiveness, our method can still achieve promising attacking performance. 

\section{Conclusion}
In this paper, we reveal the patch self-coupling issue that severely limits the robustness of existing patch-based attacks. This issue leads to the strict necessity for patch integrity and thus inherently undermines the patch robustness in complex real-world conditions. Upon this observation, we propose the Decoupled adversarial Patch (DePatch) to address the patch self-coupling issue. Specifically, we apply block-wise decoupling with the border shifting during patch optimization and introduce a variant-adjusting strategy named PDS. In the digital world, DePatch outperforms other patch-based methods under various settings of simulated physical transformations. In the more critical physical world, DePatch can be deployed by either covering clothes with expandable patches or attaching adversarial posters. Experiments on both forms of real-world application demonstrate the superior robustness of our method. 
\bibliographystyle{ACM-Reference-Format}
\bibliography{refer.bib}

\end{document}